\documentclass[runningheads]{llncs}
\usepackage[T1]{fontenc}
\usepackage{graphicx,verbatim}
\usepackage{booktabs}
\usepackage{multirow}
\usepackage{algorithm}
\usepackage{algpseudocode}   
\usepackage{amsmath}
\usepackage{amssymb}
\usepackage{hyperref}
\usepackage{adjustbox}
\usepackage{threeparttable}
\usepackage{pifont}

\begin{document}
\title{Improving Medical Image Generative Models with Fréchet Distance Loss}

\author{Andrew Marshall\inst{1} \and
Xuanang Xu\inst{2} \and
Xiaoran Zhang\inst{1} \and
Rui Wang\inst{1} \and
Lawrence Staib\inst{1,2,3} \and
James Duncan\inst{1,2,3}}
\authorrunning{A. Marshall et al.}
\institute{Department of Biomedical Engineering, Yale University, \and
Department of Radiology \& Biomedical Imaging, Yale University, \and
Department of Electrical Engineering, Yale University,\\
\email{andrew.marshall@yale.edu}}
  
\maketitle              
\begin{abstract}
Diffusion generative models have demonstrated immense potential for synthetic medical image generation. However, these models often struggle to capture complex morphological characteristics of heterogeneous tumors with irregular boundaries, limiting their utility for downstream clinical tasks such as segmentation. This limitation stems from the standard denoising objective: minimizing a per-pixel error, which smooths high-variance irregular structures characteristic of tumors. To address this, we propose finetuning these generative models with Fréchet Distance loss (FD-loss). FD-loss aligns the first and second order feature statistics of real and generated images in a pretrained encoder space, encouraging the generator to capture complex structural variations characteristic of heterogeneous tumors. We integrate FD-loss across diverse architectural settings, using both natural- and medical-image encoders on multiple liver and brain cancer datasets spanning CT and MRI modalities. Downstream segmentation networks trained on our FD-regularized synthetic data consistently achieve superior performance, improving tumor DSC by $>$$5\%$ over unregularized synthetic augmentation alone. Qualitative analysis suggests these gains are associated with more faithful tumor synthesis and fewer segmentation hallucinations. Our results show FD-loss as an effective regularizer for medical image generative models to improve clinical workflows.


\end{abstract}
\section{Introduction}
Generative models have emerged as a promising paradigm for synthesizing medical image data. Recent methods, ranging from unconditional medical image generation to segmentation-guided diffusion \cite{konz_anatomically-controllable_2024,zhang_generative_2025} and joint image-mask pair synthesis \cite{mao_medsegfactory_2025}, have shown potential for producing high-fidelity synthetic data to augment low-data regimes. However, these models often struggle to capture structures with large variations in size, shape, and other morphological characteristics~\cite{konz_anatomically-controllable_2024}. Such failures can yield unrealistic synthetic images that, when used for downstream task training (e.g., data augmentation for tumor segmentation), degrade model performance on real clinical samples.

During generative model training, image quality is commonly assessed using distributional metrics. Fréchet Inception Distance (FID) measures the similarity between image distributions using features extracted from an InceptionV3 encoder trained on natural images~\cite{heusel_gans_2018,szegedy_rethinking_2016}. However, FID does not always correlate with medical image generation quality~\cite{konz_anatomically-controllable_2024}. 
In the medical imaging domain, RadInceptionV3 replaces the natural-image encoder with an InceptionV3 model pretrained on RadImageNet, a dataset of 1.35 million annotated CT, MRI, and ultrasound exams~\cite{mei_radimagenet_2022}. This substitution enables medical images to be analyzed in a more domain-appropriate feature space with RadFID~\cite{woodland_feature_2024}. Likewise, Fréchet Radiomic Distance (FRD)~\cite{konz_frechet_2026} compares image distributions using radiomic features. Notably, segmentation accuracy on translated medical images has been shown to correlate with both RadFID~\cite{woodland_feature_2024} and FRD~\cite{konz_frechet_2026}. Given this relationship, it is natural to ask whether Fréchet distance can serve not only as an evaluation metric but also as a loss function for improving generative models designed to support segmentation.

FID has recently been introduced as a loss function for training generative models on natural images~\cite{mathiasen_backpropagating_2021,yang_representation_2026}. Previously, using FID as a loss was impractical because reliable Fréchet Distance (FD) estimation required large sample sizes, often on the order of 50,000 images. Generating this many samples during training is empirically infeasible for small medical imaging datasets. Yang et al.~\cite{yang_representation_2026} addressed this limitation with a decoupled optimization framework in which distribution statistics are precomputed from ground-truth images. They used a small queue of generated images ($n=5,000$) to provide a useful FD signal during optimization. Whereas prior work has focused on optimizing FID for general image generators, its utility for medical image generation remains unexplored. In this study, we investigate whether FD-loss can improve segmentation-guided generative models by encouraging more realistic synthetic medical images. Specifically, we finetune generative models with FD-loss computed across multiple embedding spaces defined by different pretrained image encoders, including InceptionV3, RadInceptionV3, BiomedCLIP, and MedDINOv3. We then evaluate the resulting synthetic images through downstream medical image segmentation experiments. Our contributions are threefold: 
(1) we adapt queue-based FD-loss for finetuning segmentation-guided medical diffusion models; 
(2) we evaluate multiple natural, radiological, and biomedical feature spaces for optimization; 
(3) we demonstrate on liver CT and brain MRI datasets that FD-regularized synthetic augmentation outperforms traditional strategies.

\section{Preliminaries}
\subsection{Segmentation-Guided Generative Models}
Segmentation-guided diffusion models generate medical images by conditioning the denoising process on an anatomical segmentation mask~\cite{konz_anatomically-controllable_2024}. Let $x_0 \in \mathbb{R}^{c \times h \times w}$ denote a medical image and let $m \in \{0, \ldots, C-1\}^{h \times w}$ denote a multi-class anatomical mask. Rather than modeling the unconditional image distribution $p(x_0)$, the segmentation-guided model learns the conditional distribution $p(x_0 \mid m)$ so that generated images follow the spatial anatomy specified by the input mask. The forward noising process is unchanged from a standard diffusion model, $x_t = \sqrt{\bar{\alpha}_t}\,x_0 + \sqrt{1-\bar{\alpha}_t}\,\epsilon$, with $\epsilon \sim \mathcal{N}(0, I)$ \cite{ho_denoising_2020}. Rather than predicting the added noise $\epsilon$, we adopt the velocity $v$ parameterization~\cite{salimans_progressive_2022}, in which the network regresses the velocity target $v_t = \sqrt{\bar{\alpha}_t}\,\epsilon - \sqrt{1-\bar{\alpha}_t}\,x_0$. In practice, the mask is concatenated channel-wise with the noisy image and passed to a UNet, yielding the velocity estimate $v_\theta(x_t, t \mid m)$. The model is trained with the mask-conditional objective $\mathcal{L}_m = \mathbb{E}_{(x_0,m),t,\epsilon}\big[\| v_t - v_\theta(\sqrt{\bar{\alpha}_t}\,x_0 + \sqrt{1-\bar{\alpha}_t}\,\epsilon,\, t \mid m)\|^2\big]$. Because the mask is supplied throughout the iterative denoising trajectory, the generator receives explicit spatial guidance for anatomical structures while retaining stochastic variation in image appearance. The clean image can be recovered from the predicted velocity via $\hat{x}_0 = \sqrt{\bar{\alpha}_t}\,x_t - \sqrt{1-\bar{\alpha}_t}\,v_\theta(x_t, t \mid m)$, and sampling is performed with an accelerated sampler such as DDIM~\cite{song_denoising_2022} to synthesize images anatomically aligned with the conditioning mask.

\subsection{Fréchet Distance}
Fréchet distance compares two image sets using statistics computed from their feature representations. The real image set $\mathcal{R} = \{\mathbf{x}_i\}$ and the generated image set $\mathcal{G} = \{\hat{\mathbf{x}}_i\}$ are passed through a feature extractor $\phi(\cdot)$, and the resulting feature distributions are approximated as multivariate Gaussians with means $\boldsymbol{\mu}_r = \mathbb{E}[\phi(\mathbf{x})]$, $\boldsymbol{\mu}_g = \mathbb{E}[\phi(\hat{\mathbf{x}})]$ and covariances $\boldsymbol{\Sigma}_r = \mathrm{Cov}[\phi(\mathbf{x})]$, $\boldsymbol{\Sigma}_g = \mathrm{Cov}[\phi(\hat{\mathbf{x}})]$. The distance between these two Gaussians combines the squared difference between their means with a covariance-matching term:
\begin{equation}
\label{eq:fd}
    \mathrm{FD}_\phi(\mathcal{R}, \mathcal{G})
    = \lVert \boldsymbol{\mu}_r - \boldsymbol{\mu}_g \rVert_2^2
    + \mathrm{Tr}\!\left( \boldsymbol{\Sigma}_r + \boldsymbol{\Sigma}_g
    - 2\left( \boldsymbol{\Sigma}_r \boldsymbol{\Sigma}_g \right)^{\frac{1}{2}} \right).
\end{equation}
When $\phi$ is chosen as InceptionV3~\cite{szegedy_rethinking_2016}, this formulation reduces to the familiar Fréchet Inception Distance (FID)~\cite{heusel_gans_2018}. When FD is used as an evaluation metric, the real-image statistics $(\boldsymbol{\mu}_r, \boldsymbol{\Sigma}_r)$ are computed once over the training set, whereas the generated-image statistics $(\boldsymbol{\mu}_g, \boldsymbol{\Sigma}_g)$ are estimated from a large pool of samples, often numbering in the tens of thousands.

\section{Methods}

\begin{figure}[t]
    \centering
    \includegraphics[width=0.95\linewidth]{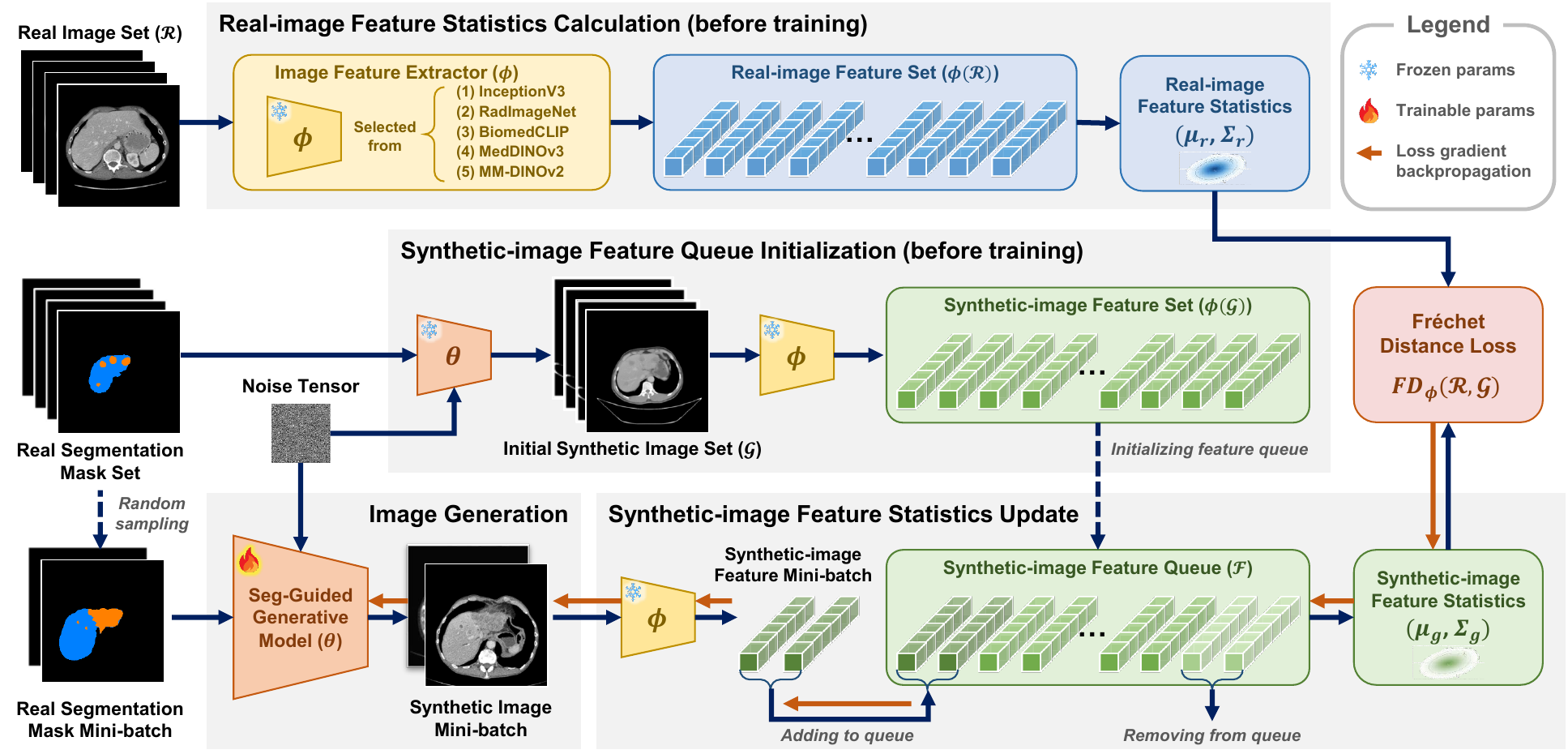}
    \caption{Our proposed segmentation-guided generative model finetuning with FD-loss.}
    \label{fig:fd_architecture}
\end{figure}

\subsection{Finetuning Generative Models with Fréchet Distance Loss}
Following the FD-loss framework~\cite{yang_representation_2026}, we finetune a pretrained segmentation-guided generator by minimizing the Fréchet distance between real and generated image distributions in a frozen representation space. For a chosen feature extractor $\phi(\cdot)$, the real-image statistics $(\boldsymbol{\mu}_r, \boldsymbol{\Sigma}_r)$ are precomputed once from the training set and held fixed during optimization. Before finetuning begins, the pretrained generator synthesizes $N=5{,}000$ images from real segmentation masks using $25$-step DDIM sampling. These images are passed through the frozen extractor to obtain $N$ synthetic-image features, which initialize a feature queue $\mathcal{F}$ of fixed size $N$ used to estimate the generated-image distribution.

During finetuning, each iteration samples a mini-batch of $M=16$ segmentation masks and synthesizes $M$ images through a differentiable $25$-step DDIM trajectory, retaining gradients through the full sampling chain. To accelerate queue turnover without incurring the memory cost of backpropagating through additional sampling trajectories, we simultaneously generate a further $32$ images per iteration without gradient tracking; these images are also passed through $\phi(\cdot)$ and added to the queue, but contribute only to the generated-image statistics, not to the loss gradient. All $48$ new features per iteration are added to $\mathcal{F}$, and the oldest $48$ features are removed, keeping the queue size fixed at $N$. The generated statistics $(\boldsymbol{\mu}_g, \boldsymbol{\Sigma}_g)$ are then computed from this updated queue, and the FD between the fixed real statistics and the queue-based generated statistics is used as the training objective (Eq.~\ref{eq:fd}). Gradients are propagated only through the $M=16$ gradient-tracked images and the generator parameters, while all other queue features remain detached. This decouples the population size needed for stable FD estimation from the batch size used for gradient computation, making FD practical as a finetuning loss for medical image generators.
We finetune each generator for $20{,}000$ steps using a learning rate of $1\mathrm{e}{-7}$, monitoring the FD-loss trajectory to confirm convergence, on a single NVIDIA H200 GPU.

\subsection{Training Segmentation Models with Synthetic Data}

To evaluate the realism and downstream utility of our synthetic data, we trained segmentation models and evaluated their performance on real clinical scans. We trained a standard 2D U-Net~\cite{ronneberger_u-net_2015} for 100 epochs using a batch size of 8, an initial learning rate of $10^{-3}$, and a cosine annealing schedule with the MONAI \cite{cardoso_monai_2022} framework. For each run, we retained the model weights that achieved the lowest validation loss. To mitigate the run-to-run variance inherent to single-seed U-Net training, we trained an ensemble of five models for each configuration. We used a fixed set of random seeds across all experiments so that every configuration was evaluated under identical initializations, ensuring that performance differences are attributable to FD-loss regularization. During inference, class probabilities predicted by the five models were averaged, and a pixel-wise argmax was applied to obtain the final segmentation mask. All downstream quantitative results are reported using this soft-voting ensemble approach.

\begin{figure}[t]
    \centering
    \includegraphics[width=0.89\linewidth]{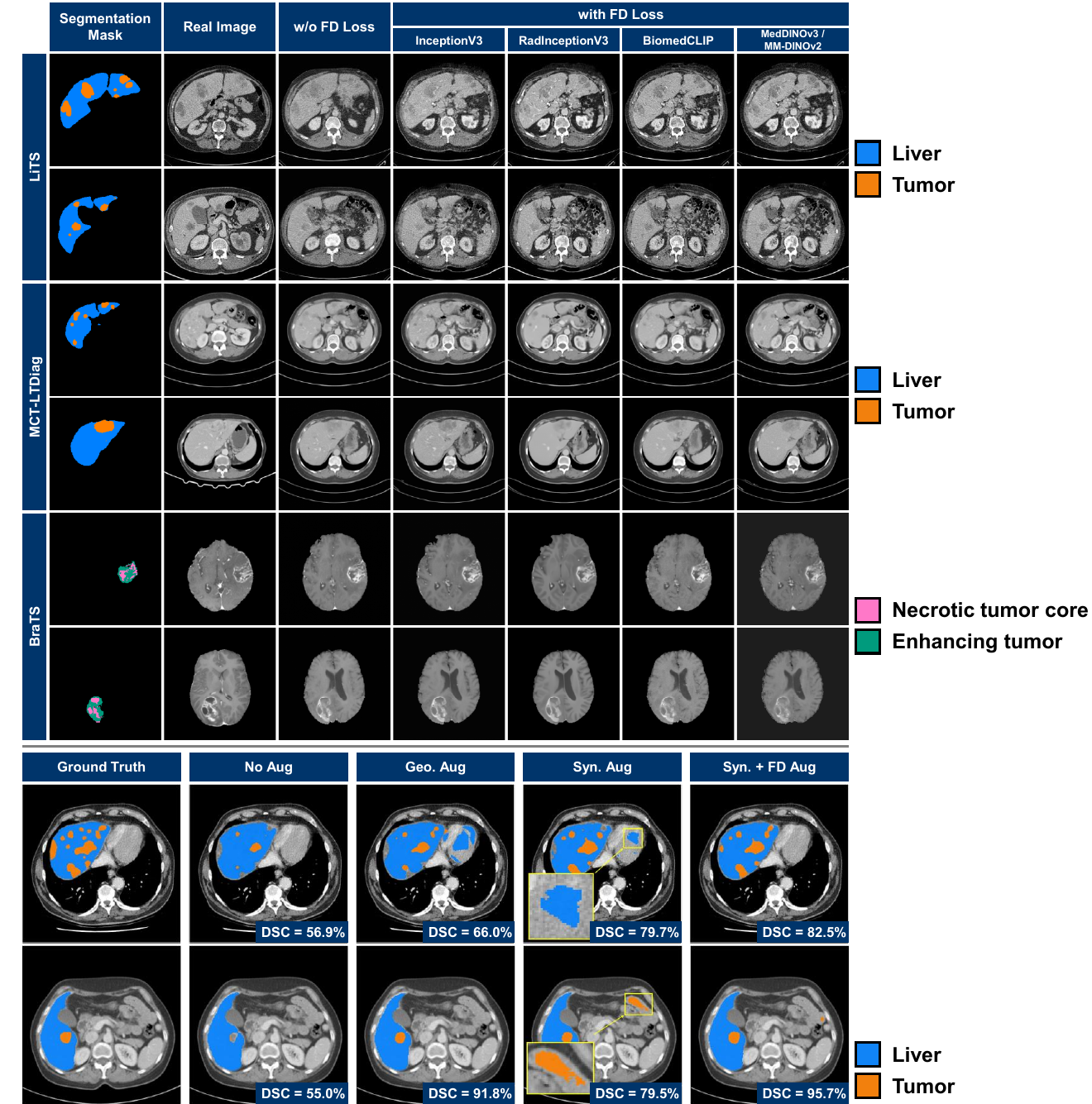}
    \caption{Qualitative evaluation of FD-loss finetuning for image generation and downstream segmentation. Top: segmentation-conditioned synthetic images generated with different encoder spaces. Bottom: downstream liver CT segmentation results.} 
    \label{fig:seg-guided-diffusion-generation}
\end{figure}

\begin{table}[t]
\centering

    \begin{minipage}[t]{0.47\textwidth}
    \centering

        \caption{Synthetic image quality before and after FD-loss finetuning. Best \textbf{bold}.}
        \label{tab:gen_quality}
        \resizebox{0.95\linewidth}{!}{%
        \begin{tabular}{lccccc}
        \toprule
        \textbf{Models}
        & \textbf{FID} $\downarrow$ & \textbf{KID} $\downarrow$ & \textbf{IS} $\uparrow$ & \textbf{CMMD} $\downarrow$ & \textbf{FRD} $\downarrow$ \\
        \midrule
        LiTS & 99.20 & 0.1020 & 1.78 & 0.5915 & 63.56 \\
        LiTS+FD & \textbf{61.37} & \textbf{0.0526} & \textbf{1.93} & \textbf{0.1861} & \textbf{24.05} \\
        \midrule
        MCT & 20.40 & 0.0090 & 1.91 & 0.0740 & 69.19 \\
        MCT+FD & \textbf{18.15} & \textbf{0.0061} & \textbf{1.95} & \textbf{0.0329} & \textbf{4.37} \\
        \midrule
        BraTS & 38.32 & \textbf{0.0294} & 1.61 & 0.3899 & 58.45 \\
        BraTS+FD & \textbf{36.67} & 0.0298 & \textbf{1.77} & \textbf{0.2683} & \textbf{5.98} \\
        \bottomrule
        \end{tabular}%
        }

        \vspace{0.4cm}

        \caption{Downstream segmentation with different data augmentations. Best results \textbf{bold}, second best \underline{underlined}.}
        \label{tab:results}
        \resizebox{0.95\linewidth}{!}{%
        \begin{threeparttable}
        \begin{tabular}{@{}c@{\hspace{3pt}}l cc cc@{}}
        \toprule
        \multirow{2}{*}{\textbf{Data}} & \multirow{2}{*}{\textbf{Aug.}}
        & \multicolumn{2}{c}{\textbf{DSC \scriptsize{[\%]}} $\uparrow$}
        & \multicolumn{2}{c}{\textbf{ASSD \scriptsize{[mm]}} $\downarrow$} \\
        \cmidrule(lr){3-4}\cmidrule(lr){5-6}
        & & \textbf{Liv.} & \textbf{Tum.} & \textbf{Liv.} & \textbf{Tum.} \\
        \midrule
        \multirow{4}{*}{\rotatebox[origin=c]{90}{\textbf{LiTS}}}
        & No aug. & \textbf{88.5} & 14.5 & \textbf{4.5} & 23.9 \\
        & Geo. & 85.9 & 44.9$^{*}$ & 13.1 & 14.2$^{*}$ \\
        & Syn. & 86.6 & \underline{46.9}$^{*}$ & \underline{10.0} & \underline{13.9}$^{*}$ \\
        & Syn.+FD & \underline{87.8} & \textbf{52.3}$^{\ddagger}$ & 10.6 & \textbf{13.7}$^{\ddagger}$ \\
        \addlinespace[2pt]\midrule\addlinespace[1pt]
        \multirow{4}{*}{\rotatebox[origin=c]{90}{\textbf{MCT}}}
        & No aug. & \underline{88.2} & 45.9 & \textbf{5.7} & 10.4 \\
        & Geo. & \textbf{88.3} & 60.8$^{*}$ & \underline{6.0} & \textbf{8.9}$^{*}$ \\ 
        & Syn. & 87.9 & \underline{60.9}$^{*}$ & 6.7 & \underline{9.8}$^{*}$ \\
        & Syn.+FD & 88.1 & \textbf{64.5}$^{\ddagger}$ & 7.4 & 9.9$^{*}$ \\
        \addlinespace[2pt]\midrule\addlinespace[1pt]
        & & \textbf{NCR} & \textbf{ET} & \textbf{NCR} & \textbf{ET} \\
        \multirow{3}{*}{\rotatebox[origin=c]{90}{\textbf{BraTS}}}
        & No aug. & 36.3 & 61.4 & 9.4 & 7.7 \\
        & Geo. & \textbf{47.6}$^{*}$ & \textbf{68.2}$^{*}$ & 8.8$^{*}$ & \underline{5.7}$^{*}$ \\
        & Syn. & 46.4$^{*}$ & 65.0$^{*}$ & \underline{8.1}$^{*}$ & 6.0$^{*}$ \\
        & Syn.+FD & \underline{47.4}$^{*}$ & \underline{66.9}$^{*}$ & \textbf{7.3}$^{\ddagger}$ & \textbf{4.9}$^{\ddagger}$ \\
        \bottomrule
        \end{tabular}
        \begin{tablenotes}
            \item[*] denotes significance vs. no augmentation; 
            \item[$^{\ddagger}$] denotes significance vs. both no augmentation and geometric augmentation.
        \end{tablenotes}
        \end{threeparttable}%
        }
    
    \end{minipage}\hfill%
    \begin{minipage}[t]{0.47\textwidth}
    \centering
    
        \caption{Feature-extractor ablation.}
        \label{tab:encoder_results}
        \resizebox{\linewidth}{!}{%
        \begin{threeparttable}
        \begin{tabular}{@{}c@{\hspace{3pt}}l cc cc@{}}
        \toprule
        \multirow{2}{*}{\textbf{Data}} & \multirow{2}{*}{\textbf{Encoder}}
        & \multicolumn{2}{c}{\textbf{DSC \scriptsize{[\%]}} $\uparrow$}
        & \multicolumn{2}{c}{\textbf{ASSD \scriptsize{[mm]}} $\downarrow$} \\
        \cmidrule(lr){3-4}\cmidrule(lr){5-6}
        & & \textbf{Liv.} & \textbf{Tum.} & \textbf{Liv.} & \textbf{Tum.} \\
        \midrule
        \multirow{4}{*}{\rotatebox[origin=c]{90}{\textbf{LiTS}}}
        & InceptionV3 & 83.5 & 44.5 & \underline{11.6} & 20.4 \\
        & RadInceptionV3 & 86.1 & 47.2 & 13.1 & 19.3 \\
        & BiomedCLIP & \underline{86.7} & \underline{47.7}$^{*}$ & 11.7 & \underline{16.6} \\
        & MedDINOv3 & \textbf{87.8}$^{*}$ & \textbf{52.3}$^{*}$ & \textbf{10.6} & \textbf{13.7}$^{*}$ \\
        \addlinespace[2pt]\midrule\addlinespace[1pt]
        \multirow{4}{*}{\rotatebox[origin=c]{90}{\textbf{MCT}}}
        & InceptionV3 & 87.5 & 55.8 & 7.5 & 10.5 \\
        & RadInceptionV3 & 82.8 & 52.4 & 10.3 & 22.2 \\
        & BiomedCLIP & \underline{87.6} & \underline{62.9}$^{*}$ & \underline{7.4} & \underline{10.5} \\
        & MedDINOv3 & \textbf{88.1} & \textbf{64.5}$^{*}$ & \textbf{7.4} & \textbf{9.9}$^{*}$ \\
        \addlinespace[2pt]\midrule\addlinespace[1pt]
        & & \textbf{NCR} & \textbf{ET} & \textbf{NCR} & \textbf{ET} \\
        \multirow{3}{*}{\rotatebox[origin=c]{90}{\textbf{BraTS}}}
        & InceptionV3 & 42.9 & 65.3 & \underline{8.0} & \underline{5.9} \\
        & RadInceptionV3 & 38.8 & 63.3 & 8.3 & 6.3 \\
        & BiomedCLIP & \textbf{47.4}$^{*}$ & \textbf{66.9}$^{*}$ & \textbf{7.3}$^{*}$ & \textbf{4.9}$^{*}$ \\
        & MM-DINOv2 & \underline{46.0} & \underline{65.5}$^{*}$ & 9.9 & 6.8 \\
        \bottomrule
        \end{tabular}
        \begin{tablenotes}
            \item[*] denotes statistical significance relative to the unregularized segmentation-guided synthetic-augmentation baseline.
        \end{tablenotes}
        \end{threeparttable}%
        }

        \vspace{0.4cm}

        \caption{Distance-function ablation.}
        \label{tab:ablation_results}
        \resizebox{0.90\linewidth}{!}{%
        \begin{threeparttable}
        \begin{tabular}{@{}l cc cc@{}}
        \toprule
        \multirow{2}{*}{\textbf{Method}}
        & \multicolumn{2}{c}{\textbf{DSC \scriptsize{[\%]}} $\uparrow$}
        & \multicolumn{2}{c}{\textbf{ASSD \scriptsize{[mm]}} $\downarrow$} \\
        \cmidrule(lr){2-3}\cmidrule(lr){4-5}
        & \textbf{Liv.} & \textbf{Tum.}
        & \textbf{Liv.} & \textbf{Tum.} \\
        \midrule
        FD $\mu$ Only
        & 81.8$^{*}$ & 27.3$^{*}$ & 14.4$^{*}$ & 27.1$^{*}$ \\
        FD $\Sigma$ Only
        & 86.7$^{*}$ & 41.7$^{*}$ & 11.9$^{*}$ & 18.5$^{*}$ \\
        Symmetric KL
        & \underline{86.9}$^{*}$ & 38.8$^{*}$ & 11.6$^{*}$ & 21.2$^{*}$ \\
        Bhattacharyya
        & 86.8$^{*}$ & \underline{44.2}$^{*}$ & \textbf{10.5} & \underline{17.1}$^{*}$ \\
        \midrule
        FD
        & \textbf{87.8} & \textbf{52.3} & \underline{10.6} & \textbf{13.7} \\
        \bottomrule
        \end{tabular}
        \begin{tablenotes}
            \item[*] denotes statistical significance relative to FD baseline.
        \end{tablenotes}
        \end{threeparttable}%
        }
    
    \end{minipage}
\end{table}

\section{Experiments}
\subsection{Datasets and Metrics}
We evaluate our method on two contrast-enhanced liver CT datasets and one multi-parametric brain MRI dataset. \textbf{Liver Tumor Segmentation (LiTS)}~\cite{bilic_liver_2023} CT dataset contains both primary and secondary hepatic tumors. Following the experimental design of Konz et al.~\cite{konz_anatomically-controllable_2024}, we partition the data at the patient level into three mutually exclusive subsets: 100 patients for generative model training ($\sim$13,250 images), 16 patients for generative testing ($\sim$3,000 images), and an additional held-out set of 15 patients ($\sim$2,600 images) reserved for downstream segmentation testing. \textbf{Multi-phase CT dataset for Liver Tumor Diagnosis (MCT-LTDiag)}~\cite{wu_multi-phase_2025} comprises 517 patients spanning HCC, ICC, and colorectal and breast-cancer liver metastases. We partition this dataset into 300 patients ($\sim$12,890 images) for generative model training, 110 patients ($\sim$2,000 images) for generative testing, and 105 patients ($\sim$2,000 images) for downstream segmentation testing. For both CT datasets, volumes are windowed to a Hounsfield unit (HU) range of $[-200, 250]$, and axial slices are saved at $256\times256$ pixels and normalized to $[0,1]$. To assess generalization beyond abdominal CT, we further evaluate on the \textbf{Brain Tumor Segmentation (BraTS)} dataset~\cite{menze_multimodal_2015,bakas_advancing_2017,bakas_identifying_2019} containing multi-parametric brain MRI with glioma sub-regions.  For the BraTS dataset 260 patients ($\sim$11,841 images) were used for generative model training, 59 ($\sim$2,598 images) were used for generative testing, and 50  ($\sim$2,305 images) were used for downstream task evaluation.
We generate T1Gd images conditioned on enhancing tumor (ET) and necrotic core (NCR) labels.

We evaluate synthetic image quality using distributional metrics that compare generated and real images: Fréchet Inception Distance (FID) \cite{heusel_gans_2018}, Kernel Inception Distance (KID) \cite{binkowski_demystifying_2021}, Inception Score (IS) \cite{salimans_improved_2016}, CLIP Maximum Mean Discrepancy (CMMD) \cite{jayasumana_rethinking_2024}, and Fréchet Radiomic Distance (FRD) \cite{konz_frechet_2026}. Downstream segmentation utility is assessed by training segmentation models under each synthetic-data configuration and testing them on held-out real images using the Dice similarity coefficient (DSC) and average symmetric surface distance (ASSD).
Statistical significance between augmentation configurations was assessed using a Wilcoxon signed-rank test. Significance is reported at $p$<0.05.

\subsection{Synthetic Image Quality Evaluation}
We first evaluate whether FD-loss directly improves generative model performance by enhancing synthetic image quality. For each dataset, we compare synthetic images produced by the baseline segmentation-guided generative model trained without FD-loss with images produced by the same model finetuned with FD-loss. Both models are sampled using the same conditioning masks, and the resulting synthetic image distributions are compared with the corresponding real-image distributions using the image-quality metrics described above (MedDINOv3 for liver CT, BiomedCLIP for BraTS; see Table~\ref{tab:encoder_results}). The quantitative results in Table~\ref{tab:gen_quality} demonstrate the effectiveness of FD-loss: after finetuning, generated samples generally move closer to the real data distribution across LiTS, MCT-LTDiag, and BraTS, as reflected by lower FID, KID, CMMD, and FRD scores and an IS closer to the real-image reference. Representative qualitative examples are shown in Fig.~\ref{fig:seg-guided-diffusion-generation}.

\subsection{Downstream Segmentation with Synthetic Data Augmentation}
We next evaluate whether FD-regularized generative models improve downstream segmentation when used as an advanced data augmentation strategy. This experiment provides more direct evidence for the significance of FD-loss in generative model finetuning, because the primary purpose of synthetic images is to improve downstream model training rather than merely to improve image-level metrics. If segmentation models trained with FD-regularized synthetic data outperform models trained with conventional augmentation alone, this demonstrates both the effectiveness and practical value of the proposed approach. We therefore train segmentation models under four progressively stronger settings: (1) no data augmentation; (2) conventional augmentation only, including random scaling, rotation, flipping, deformation, and intensity rescaling; (3) augmentation with synthetic images generated by the baseline model trained without FD-loss; and (4) augmentation with synthetic images generated by the model finetuned with FD-loss. The results in Table~\ref{tab:results} show that FD-regularized synthetic images (``\textbf{Syn.+FD}'') generally improve segmentation performance relative to the unaugmented (``\textbf{No aug.}''), conventionally augmented (``\textbf{Geo.}''), and unregularized synthetic-data (``\textbf{Syn.}'') baselines, demonstrating the practical utility of FD-loss for downstream segmentation training.

\subsection{Ablation Study}
\textbf{Image Feature Extractor}
We further perform an encoder-selection experiment to justify the choice of image feature extractor $\phi(\cdot)$ in the FD-loss framework (Fig.~\ref{fig:fd_architecture}). Because FD-loss optimizes the generator in the representation space defined by $\phi(\cdot)$, different feature extractors may emphasize different anatomical, textural, or modality-specific properties of the generated images. We therefore finetune the generative models using FD-loss computed with multiple candidate feature extractors.
Specifically, we consider InceptionV3~\cite{szegedy_rethinking_2016}, a CNN trained on 1.2 million natural images across 1,000 classes; 
RadInceptionV3~\cite{mei_radimagenet_2022}, an InceptionV3 model trained on 1.35 million CT, MRI, and ultrasound images spanning neurological, oncological, and abdominal imaging; BiomedCLIP~\cite{zhang_biomedclip_2025}, a ViT trained on 15 million biomedical image-text pairs from 4.4 million scientific articles; MedDINOv3~\cite{li_meddinov3_2025}, a ViT trained on 3.9 million 2D abdominal CT images; 
and MM-DINOv2~\cite{scholz_mm-dinov2_2025}, a ViT trained on 50,000 brain MRI volumes. 
As shown in Table~\ref{tab:encoder_results}, the optimal representation is dataset dependent: MedDINOv3 achieves the best overall performance on the abdominal CT datasets, LiTS and MCT-LTDiag, whereas BiomedCLIP performs best on the BraTS brain MRI dataset. These results support the use of domain- and modality-appropriate feature extractors when applying FD-loss to medical image generation.

\textbf{Distance Function} 
To validate the choice of Fréchet Distance, we perform an ablation study substituting alternative distance functions. We investigate the impact of only tracking the first or second statistical moment, and using alternative distance functions for probability distributions such as symmetric KL divergence and Bhattacharyya distance. Calculating both symmetric KL divergence and Bhattacharyya distance require inverting $\Sigma$, which could introduce instability into training. To avoid this, we use Tikhonov regularization: $\Sigma_{stable} = \Sigma + \epsilon I$; we select $\epsilon = 10^{-6}$. We test each distance function with the LiTS dataset \cite{bilic_liver_2023} in the MedDINOv3 representation space \cite{li_meddinov3_2025}. The results of the ablation study are reported in Table~\ref{tab:ablation_results}, which showed that utilizing FD yields the best downstream task performance.

\section{Conclusion}
In this work, we evaluated Fréchet Distance loss as a finetuning objective for segmentation-guided medical image generative models. Across abdominal CT and brain MRI datasets, FD-loss improved synthetic image realism and increased downstream segmentation performance, with the largest gains observed for heterogeneous tumor regions. Our encoder study showed that the representation space used for FD-loss is an important design choice, with medical and biomedical encoders generally outperforming natural-image features while varying in effectiveness across datasets and modalities. These results suggest that feature-distribution matching is a useful regularizer for generating synthetic data that better supports medical image segmentation. Future work should extend this framework to 3D image generation and other clinical tasks, including domain harmonization and image-to-image translation.

%
%

\bibliographystyle{splncs04}
\bibliography{references}

\end{document}